\documentclass[journal]{IEEEtran}
\ifCLASSINFOpdf
\else
\fi
\usepackage{graphicx}
\usepackage{subfigure}
\usepackage{amsmath}
\usepackage[noadjust]{cite}
\usepackage{xcolor}
\usepackage{float}
\usepackage{cite}
\usepackage{enumerate}
\usepackage{cases}
\usepackage{multirow}
\usepackage{bm}
\usepackage{booktabs}
\usepackage{array}
\usepackage[ruled]{algorithm2e}
\usepackage{makecell}

\begin{document}
\title{Novel View Synthesis from a Single Image via Unsupervised Learning}

\author{Bingzheng~Liu,
        Jianjun~Lei,~\IEEEmembership{Senior Member,~IEEE,}
        Bo~Peng,~\IEEEmembership{Member,~IEEE,}
        Chuanbo~Yu,\\
        Wanqing~Li,~\IEEEmembership{Senior Member,~IEEE,}
        Nam Ling,~\IEEEmembership{Fellow,~IEEE}  % stops a space
\thanks{The work of Jianjun Lei and Bo Peng was supported in part by the National Natural Science Foundation of China (No.62125110, 62101379, 61931014),National Key R\&D Program of China (No.2018YFE0203900), and Natural Science Foundation of Tianjin (No.18JCJQJC45800). (Corresponding author: Jianjun Lei.)\emph{}}
\thanks{B. Liu, J. Lei, B. Peng,  and C. Yu are with the School of Electrical and Information Engineering, Tianjin University, Tianjin 300072, China (e-mail: bzliu@tju.edu.cn; jjlei@tju.edu.cn; bpeng@tju.edu.cn; cbyu@tju.edu.cn).} % stops a space
\thanks{W. Li is with the Advanced Multimedia Research Lab, University of Wollongong, Wollongong, Australia (e-mail: wanqing@uow.edu.au).}%
\thanks{N. Ling is with the Department of Computer Science and Engineering, Santa Clara University, Santa Clara, CA 95053 USA (e-mail: nling@scu.edu).} % stops a space
%\thanks{Q. Huang is with the School of Computer Science and Technology, University of Chinese Academy of Sciences, Beijing 101408, China (e-mail: qmhuang@ucas.ac.cn).}% <-this % stops a space
\thanks{Digital Object Identifier}}

% The paper headers
\markboth{}%
{Shell \MakeLowercase{\textit{et al.}}: Bare Demo of IEEEtran.cls for IEEE Journals}
% make the title area

\maketitle

\begin{abstract}
View synthesis aims to generate novel views from one or more given source views. Although existing methods have achieved promising performance, they usually require paired views of different poses to learn a pixel transformation. This paper proposes an unsupervised network to learn such a pixel transformation from a single source viewpoint. In particular, the network consists of a token transformation module (TTM) that facilities the transformation of the features extracted from a source viewpoint image into an intrinsic representation with respect to a pre-defined reference pose and a view generation module (VGM) that synthesizes an arbitrary view from the representation. The learned transformation allows us to synthesize a novel view from any single source viewpoint image of unknown pose. Experiments on the widely used view synthesis datasets have demonstrated that the proposed network is able to produce comparable results to the state-of-the-art methods despite the fact that learning is unsupervised and only a single source viewpoint image is required for generating a novel view. The code will be available soon.
\end{abstract}


\IEEEpeerreviewmaketitle

\section{Introduction}
\IEEEPARstart{N}{ovel} view synthesis (NVS) aims to generate an unknown-view from a single or multiple source views. Many methods have been developed to synthesize a novel view from multiple views \cite{1}-\cite{2}. Recently, methods are also explored to synthesize a novel view from a single source view \cite{3}-\cite{5}. The key underlying mechanism of these methods for synthesis from a single view is to learn a view transformation, either 2D or 3D, between a source view and a target view. Such a transformation is often learned from paired views in which one view is treated as a target view to serve as a supervising signal and the other view is considered as the source from which the target view is synthesized. The learned transformation allows us to synthesize a novel view from a single source view of known pose. However, camera pose information of the single source view must be provided for the synthesis of a novel view. In other words, only the views with pose information can be chosen as input in synthesis.
\begin{figure}[!]
\centering
\includegraphics[width=1\linewidth]{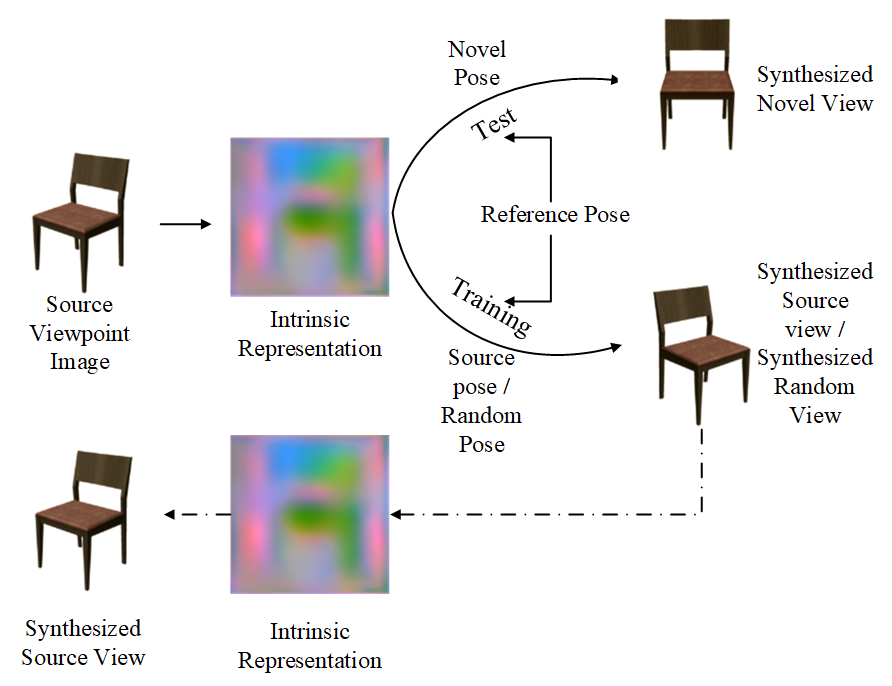}
\setlength{\abovecaptionskip}{-0.25cm}
\caption{The pipeline of the proposed method. The training are divided into two stages, and only a single view is required in training. In the first stage, the feature extracted from a source viewpoint image is transformed into the intrinsic representation with respect to a pre-defined reference pose, and the view transformation between reference pose and a source pose is learned. In the second stage, the view transformation between reference pose and a random pose is first learned, and then a reverse mapping strategy is introduced to further fine-tune the network. In synthesis, an intrinsic representation is obtained from a single viewpoint image without need for knowing its pose and a novel view of an arbitrary pose is synthesized from the intrinsic representation.}
\label{figure0}
\end{figure}

In a practical multi-view scenario \cite{6}-\cite{7}, such as broadcasting of a sports event, multiple source views are captured by a set of fixed source cameras with known poses. At the same time, there are also a few moving cameras in the scene that dynamically follows the important part of the event. It is a desirable and appealing feature if a novel view can be generated from the views taken by a moving camera or a hand-held camera in the scene. Since it is usually difficult to obtain the pose information of these moving cameras or hand-held cameras in real-time, existing methods for novel view synthesis from a single view are not applicable because they must be provided with the pose information of the input single source view.

To address this limitation, this paper proposes an unsupervised network that is able to synthesize a novel view from a single source viewpoint image without requiring the pose information of the source view. The key idea is to learn a view transformation between a pose and a pre-defined reference pose. To this end, the proposed network mainly consists (a) a specially designed token transformation module (TTM) that maps the features of any input source viewpoint image (with unknown pose information) to an intrinsic representation with respect to a reference pose, (b) a view generation module (VGM) that reconstructs an explicit occupancy volume with respect to the reference pose, rotates the volume explicitly to a target pose to generate the target view. The network is trained in an unsupervised manner. In particular, a reverse mapping strategy is introduced to improve the training. Compared to the existing methods for synthesizing novel views from a single view, the proposed unsupervised network has two advantages. First, it only requires a source viewpoint image without pose information during inference for view synthesis. Second, the network is trained using a single view, rather than paired views with different poses as most existing methods do. The pipeline of the proposed method is shown in Fig.~\ref{figure0}.

In summary, the main contributions of this paper include:

1) A new unsupervised network is proposed for novel view synthesis from a single image. Unlike existing methods, it does not require pose information of the single source view during synthesis. Therefore, choice of the single input viewpoint image in synthesis is not limited to the views captured by fixed source cameras and it can be an arbitrarily viewpoint image captured by a non-source camera.

2) A token transformation module is developed to learn an intrinsic representation and a view generation module is developed to synthesize novel views from the intrinsic representation.

3) A two-stage unsupervised  training is proposed in which the network is first trained using individual view and then fined-tuned with a reverse mapping strategy as detailed in Section II-D.

4) Experiments compared with state-of-the-art methods on both synthetic and real datasets have demonstrated the effectiveness of the proposed network.
\begin{figure*}[t]
\centering
\includegraphics[width=1\linewidth]{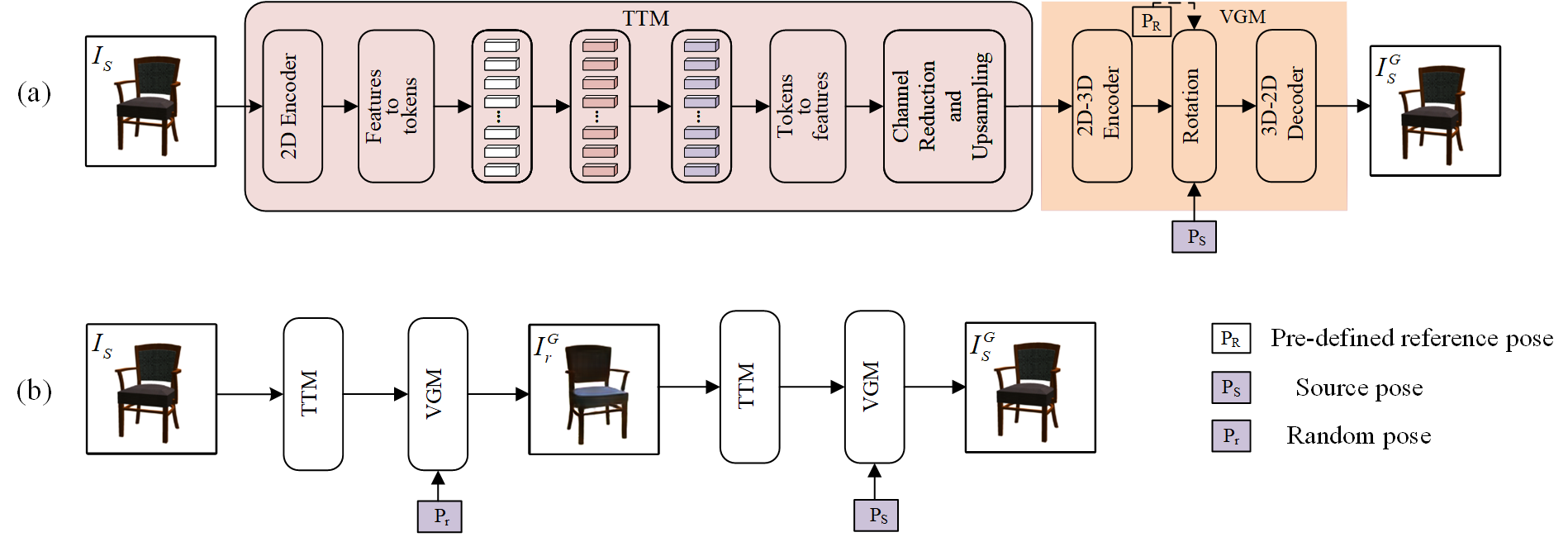}
\setlength{\abovecaptionskip}{-0.5cm}
\caption{The proposed view synthesis network. The training phase are divided into two stages. As shown in Fig. 2 (a), TTM and VGM are trained in the first stage using the source view and corresponding pose ${P}_{S}$; In the second stage, the network is fine-tuned with a synthesized view at a random pose ${P}_{r}$ through a reverse mapping strategy as shown in Fig. 2 (b); Novel views are synthesized from a source viewpoint image (without pose information) during test phase.}
\label{figure1}
\end{figure*}

The rest of this paper is organized as follows. Section II reviews the related works. Section III introduces the detail of the proposed method. The experimental results and analysis are presented in Section IV. Finally, Section V concludes this paper.

\section{Related Work}
In novel view synthesis from input source view(s), a transformation from source view to target view can be either 2D or 3D. 2D transformation-based methods mainly focus on learning pixel displacement between the input source view(s) and the target view \cite{8}-\cite{10} or directly regressing the pixel colors of the target view in its 2D image plane \cite{11}-\cite{13}. 3D transformation-based methods \cite{14}-\cite{15} often predict a 3D representation, such as an occupancy volume, first and then explicitly perform 3D spatial transformation on the representation to synthesize the target view.

\subsection{2D Transformation-Based Methods}
2D transformation-based methods are divided into two categories, namely pixels generation and prediction of pixels displacement or flow. The methods of pixels generation directly generate pixel values of a target view by using pixels regression.  Liu \emph{et al.} \cite{4} proposed a method to utilize the estimated multiple homographies between paired views through a geometric transform network. Park \emph{et al.} \cite{5} predicted disoccluded regions from input source view, and refined disoccluded regions in the synthesized target view by a transformation-based view synthesis network. Eslami \emph{et al.} \cite{12} designed a method to infer a 2D representation from a source view to generate a novel view. Tatarchenko \emph{et al.} \cite{13} proposed to directly regress pixel colors of a target view from an input source view. Alexey \emph{et al.} \cite{16} proposed a network to generate a novel view when giving the type, pose, and color of an object. Tran \emph{et al.} \cite{17} proposed  to learn a disentangled representation for pose-invariant image synthesis. Xu \emph{et al.} \cite{18} presented a view independent network to infer a target view by obtaining the essential knowledge of intrinsic properties from the object. Kusam \emph{et al.} \cite{19} presented a generative adversarial network to transfer an input source image to a target image under some conditions.

Several depth and flow prediction methods have been proposed to synthesize a novel view from a source view. For instance, Habtegebrial \emph{et al.} \cite{20} introduced a geometry estimation and view inpainting network to synthesize a novel view by utilizing the depth map predicted from a source view. In \cite{21}, a fully automated network is presented to synthesize a target view by utilizing the predicted disparity maps between paired views. In \cite{22}, an appearance flow prediction network was proposed to synthesize a novel view. Although these methods can obtain promising performance in NVS, they usually require paired views with different poses for training and a source view with known pose for synthesis.
\subsection{3D Transformation-Based Methods}
Different from the 2D transformation-based methods, 3D transformation-based methods \cite{23}-\cite{28} generate a novel view with the assistance of an estimated 3D representation from the input source view(s). Some of them obtain an explicit 3D representation with the supervision of 3D information, such as voxels \cite{29}, point-clouds \cite{30} and mesh models \cite{31}, and then the 3D representation can be rendered to a 2D view from a specified camera pose. For example, Thu \emph{et al.} \cite{25} utilized a deep differentiable convolutional network to render a view from a 3D shape of objects. In \cite{26}, a 3D recurrent reconstruction neural network was presented to obtain the 3D occupancy grid of an object. Paul \emph{et al.} \cite{29} predicted 3D shape samples from input views by jointly considering the mesh parameterizations and shading cues.

Methods have also reported to generate a 3D representation just from paired views with different poses. Rematas \emph{et al.} \cite{32} introduced a 2D-to-3D alignment method to perform a view transformation between a source view and a target view. Tulsiani \emph{et al.} \cite{33} presented a layer-structured 3D representation method for novel view synthesis. Liu \emph{et al.} \cite{34} learned the correlation among different views with respect to a predicted 3D representation via a viewer-centered network. Kyle \emph{et al.} \cite{2} focused on transforming 3D features from source pose to target pose, and then the transformed 3D feature was projected into a 2D plane for obtaining the target view.

The proposed method in this paper is a hybrid one to take the advantages of both 2D and 3D transformation. 2D transformation is learned to transform 2D features from one pose to a reference pose and 3D transformation is employed in generating a novel view from a 3D representation associated with the reference pose.

\section{The Proposed Method}
\subsection{Overview}
Fig.~\ref{figure1} shows the network architecture of the proposed method. It consists of a 2D encoder, a token transformation module (TTM),and a view generation module (VGM). The 2D encoder consisting of multiple convolutional layers extracts the features of the input single viewpoint image. The TTM learns an intrinsic representation of the input viewpoint image with respect to a pre-defined reference pose ${{P}_{R}}$. The VGM takes the intrinsic representation as input and reconstructs an explicit occupancy volume via a 2D-3D encoder. A target view is generated through 3D geometric transformation (i.e. rotation) of the occupancy volume via a 3D-2D decoder. In training, only individual source view is used as the supervised signal, and a viewpoint image at a novel pose is synthesized from a single source viewpoint image. Unlike the existing methods for synthesis of a novel view from a single source view that usually require pairwise views with different poses in training and a source view with known pose in synthesis, the proposed network is trained by using a single viewpoint image, and a novel view is synthesized from a single and arbitrary viewpoint image of an unknown pose. In addition, a reverse mapping strategy is also introduced to utilize the synthesized view at a random pose to synthesize source view by constructing the inverse mapping during training phase.

\subsection{Token Transformation Module (TTM)}
The transformation between views is required when synthesizing a novel view. To synthesize a novel view from a single input source viewpoint image without pose information, an intuitive idea is that the network generates an intrinsic representation with respect to a fixed pose from the input image, and the view transformation between pairwise poses (i.e. an arbitrary pose and reference pose) can be achieved. To this end, using only one source viewpoint image as input, a pre-defined reference pose ${P}_{R}$ is introduced to guide the learning of intrinsic representation from the input image, and achieve the transformation between a source view and a target view.

%Especially, TTM transforms the features between different viewpoint image with tokens transformation from features.
The purpose of the TTM is to transform the features extracted from a source viewpoint image into an intrinsic representation as if they are extracted from a reference pose ${P}_{R}$. The TTM first converts the features of the input source view into multiple tokens with the size of 1*1 through the features-to-tokens operation, in which each token represents the contents of a channel. These tokens are then transformed from an arbitrary pose to the reference pose via multiple linear mappings using 1-D convolutions. The transformed tokens are converted to the feature of the reference pose by an tokens-to-features operation. TTM outputs a feature map of the same spatial resolution as the input image through the channel reduction and upsampling. In this way, TTM not only facilitates the transformation, but also avoids a trivial solution when the network is trained using single view instead of paired views like most existing methods.

\subsection{View Generation Module (VGM)}
Inspired by the concept of mental rotation \cite{35}, an unseen novel view is obtained by rotating 3D objects mentally and projecting the ``mental" 3D representation into a specific pose. Therefore, when the intrinsic representation with respect to the reference pose ${P}_{R}$ is obtained, an occupancy volume characterizing the 3D information is constructed to explicitly perform the transformation between the reference pose and an arbitrary pose in 3D space. The viewpoint image of an arbitrary pose is rendered by projecting occupancy volume into a 2D space.

The VGM takes the transformed feature map as input and reconstructs an explicit occupancy volume with respect to the viewpoint image of reference pose ${P}_{R}$ through a multistage 2D and 3D encoder. An explicit 3D rotation is applied to transform the volume from the reference pose ${{P}_{R}}$ to the pose ${{P}_{S}}$ of source view during training while to the novel pose during synthesis. A synthesized view and its segment map of the specified pose are generated from the rotated occupancy volume via multistage 3D and 2D decoder. All 2D-3D encoder, rotation and 3D-2D decoder follows the same architectures as those used in \cite{2}.

%\subsection{Iterative Feedback}

%By jointly utilizing the TTM and VGM, a novel view can be synthesized by utilizing the camera pose of the novel view as ${{P}_{T}}$ to the during the test phase. However, since the purpose of novel view synthesis is to synthesize the view at a specified camera pose that is different from the input view, the novel view synthesized by trained model contains some wrong pixels induced by the pose difference between the output view and input view. Thus, there is an urgent need to construct view transformation between different camera poses in an unsupervised manner. In other words, the camera pose of the output view should be different from the camera pose the input view during the training phase. To this end, the CVSM is proposed to utilize the input view and a synthesized view at a random sampled camera pose obtained from the input view to construct paired views for the learning of the novel view synthesis. In the proposed CVSM, the weights are shared with the TTM and VGM to jointly train the novel view synthesis in an unsupervised manner.

\begin{table*}[t]
\newcommand{\tabincell}[2]{\begin{tabular}{@{}#1@{}}#2\end{tabular}}
\renewcommand\arraystretch{1.3}
\centering
\small
\caption{\label{table1} Quality of synthesized novel views on Chair and Car categories. ${L}_{1}$ distance (low is better) and SSIM (high is better) comparison.}
\begin{tabular}{p{1.5cm}<{\centering}p{2.5cm}<{\centering}p{3cm}<{\centering}p{3cm}<{\centering}p{2cm}<{\centering}p{2cm}<{\centering}p{2cm}<{\centering}}
\toprule[1.5pt]
\multirow{2}[2]{*}{Method} & \multirow{2}[2]{*}{\tabincell{c}{Whether need ${P}_{S}$ \\ during inference}}  & \multirow{2}[2]{*}{\tabincell{c}{Whether add noise to ${P}_{S}$ \\ during inference}} & \multicolumn{2}{c}{Chair} & \multicolumn{2}{c}{Car} \\
     \multicolumn{1}{c}{} & \multicolumn{1}{c}{} & \multicolumn{1}{c}{} & \multicolumn{1}{c}{\multirow{1}[1]{*}{${{L}_{1}}$}} & \multicolumn{1}{c}{\multirow{1}[1]{*}{SSIM}} & \multicolumn{1}{c}{\multirow{1}[1]{*}{${{L}_{1}}$}} & \multicolumn{1}{c}{\multirow{1}[1]{*}{SSIM}} \\
     \midrule
     STM \cite{13}  & $\surd$      & $\times$      & 0.269  & 0.870   & 0.133  & 0.911 \\
     AFF \cite{22}  & $\surd$      & $\times$      & 0.255  & 0.871  & 0.146  & 0.899 \\
     MTN \cite{1} & $\surd$      & $\times$      & 0.181  & 0.895  & 0.098  & 0.923 \\
     TBN \cite{2}  & $\surd$      & $\times$      & 0.178  & 0.895  & 0.091  & 0.927 \\
     Ours   & $\times$      & -      & \textbf{0.164}  & 0.879   & 0.143   & 0.890 \\
     \midrule
     STM \cite{13}  & $\surd$      & $\surd$      & 0.321  & 0.855  & 0.203  & 0.890 \\
     AFF \cite{22}  & $\surd$      & $\surd$      & 0.297  & 0.860   & 0.201  & 0.881 \\
     MTN \cite{1} & $\surd$      & $\surd$      & 0.295  & 0.861  & 0.226  & 0.879 \\
     \textbf{Ours}   & $\times$      & -      & \textbf{0.164}  & \textbf{0.879}   & \textbf{0.143}   & \textbf{0.890} \\
\bottomrule[1.2pt]
\end{tabular}
\end{table*}

\begin{figure*}[t]
\centering
\includegraphics[width=0.99\linewidth]{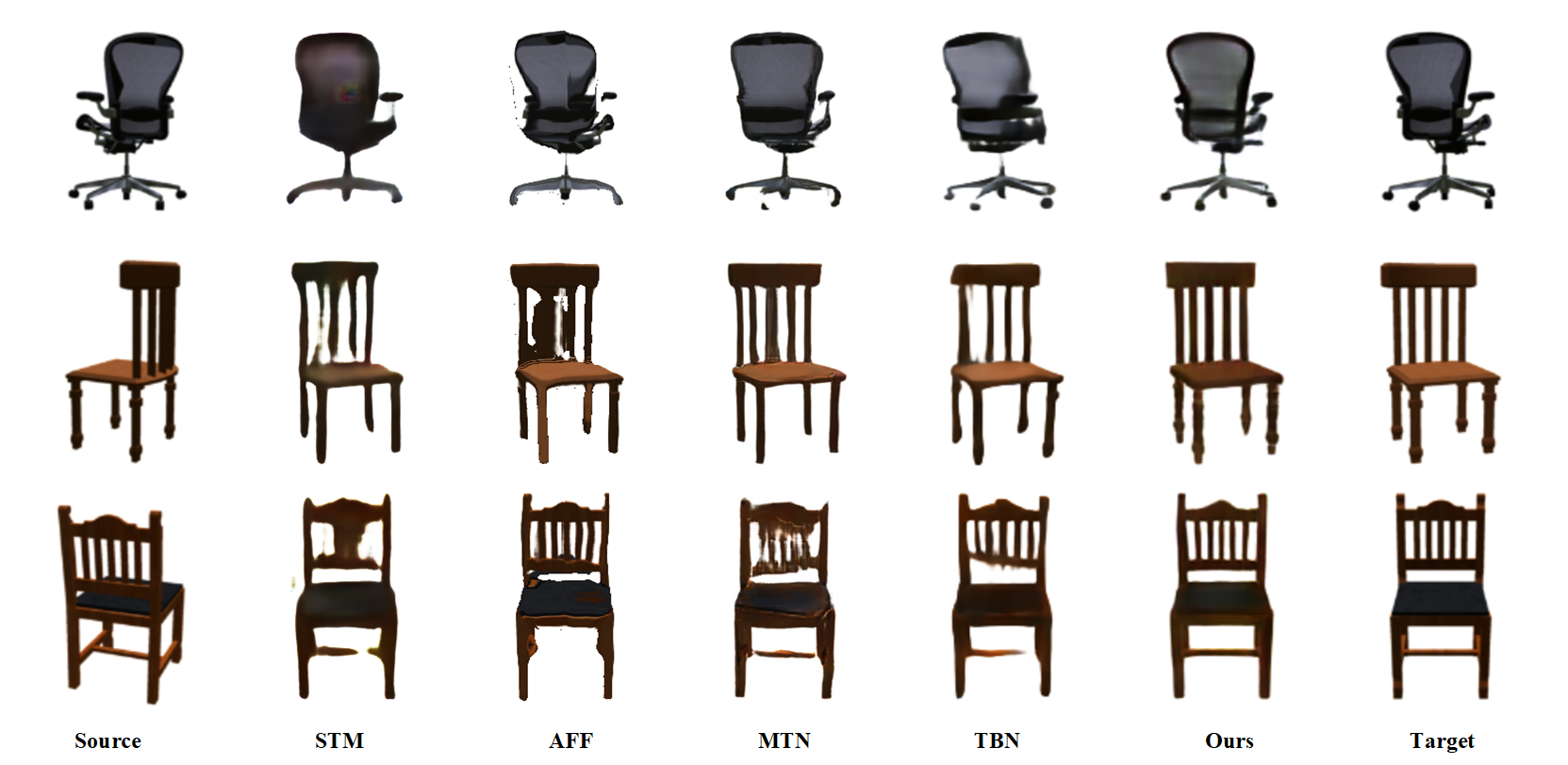}
\setlength{\abovecaptionskip}{-0.25cm}
\caption{Visual comparison of novel views of object Chair.}
\label{figure2}
\end{figure*}

\begin{figure*}[t]
\centering
\includegraphics[width=1\linewidth]{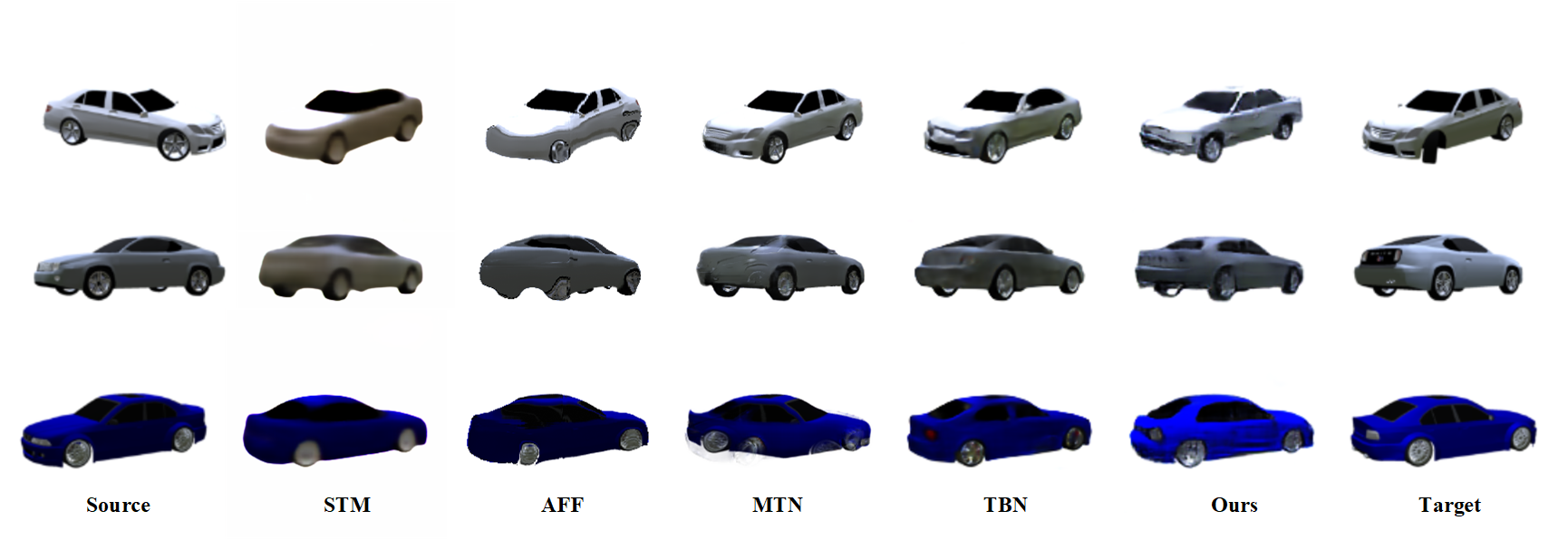}
\setlength{\abovecaptionskip}{-0.25cm}
\caption{Visual comparison of novel views of object Car.}
\label{figure3}
\end{figure*}

\begin{figure*}[t]
\centering
\includegraphics[width=0.9\linewidth]{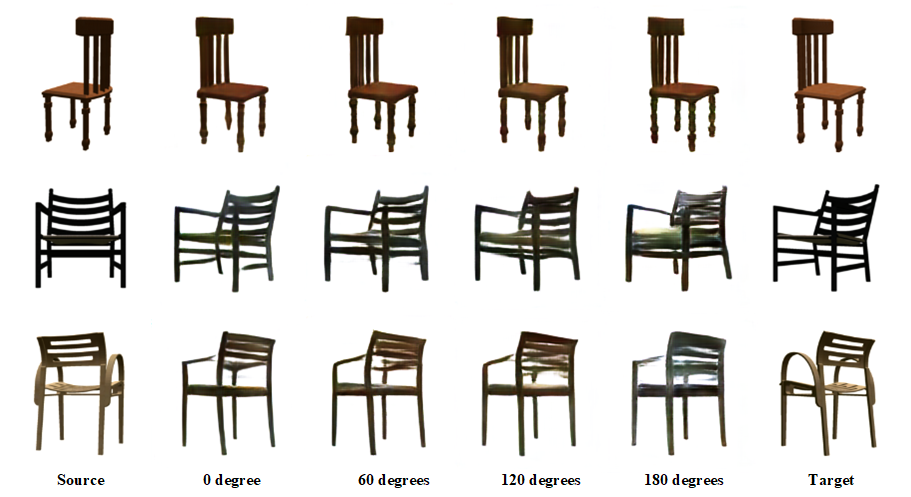}
\setlength{\abovecaptionskip}{-0.25cm}
\caption{Synthesized novel views from different azimuth angles.}
\label{figure4}
\end{figure*}

\subsection{Multi-stage Training Strategy}

In the first stage, as illustrated in Fig. 2 (a), a source view ${I}_{S}$ is input to TTM and VGM, and a view ${I}_{S}^{G}$ at the pose ${{P}_{S}}$ is synthesized. Then, the proposed network is trained by minimizing multiple losses in order to synthesize high-quality novel views, including color loss, feature loss, adversarial loss, structural similarity loss and shape loss.
Color loss is measured as a pixel-wise ${{L}_{1}}$ distance between the synthesized view and its ground-truth.
\begin{equation}\begin{array}{c}
{L}_{R}=\|{{I}_{S}} - {{I}_{S}^{G}}\|_{1}
\end{array}\label{1}\end{equation}
where ${{I}_{S}}$ is the source viewpoint image from which the target view ${{I}_{S}^{G}}$ is synthesized by VGM.

Feature loss is measured as the ${{L}_{2}}$ distance of features extracted using a pre-trained VGG-19 network $V(.)$.

\begin{equation}\begin{array}{c}
{L}_{V}=\|V({{I}_{S}}) - V({{I}_{S}^{G}})\|_{2}
\end{array}\label{2}\end{equation}

Structural similarity loss ${{L}_{SSIM}}$ \cite{36} is measured using structural similarity index measure (SSIM). Additionally, shape loss is measured by ${{L}_{1}}$ distance between the segment map of the input source viewpoint image and synthesized view.

\begin{equation}\begin{array}{c}
{L}_{S}=\|{{S}_{S}} - {{S}_{S}^{G}}\|_{1}
\end{array}\label{3}\end{equation}
where ${{S}_{S}}$ and ${{S}_{S}^{G}}$ represent the segment maps of ${{I}_{S}}$ and ${{I}_{S}^{G}}$, respectively. The segment map is calculated in the same way as that used in \cite{2}, which is an edge map.
To further improve the naturalness of the synthesized view, adversarial loss ${{L}_{A}}$ \cite{37} is also included. The total loss ${{L}_{Total}}$ is a weighted combination of the losses discussed above, that is,
\begin{equation}\begin{array}{c}
{L}_{Total} = {L}_{R} + \alpha{{L}_{SSIM}} + \beta{L}_{V} + \gamma{L}_{S} + \lambda{L}_{A}
\end{array}\label{4}\end{equation}
where $\alpha$, $\beta$, $\gamma$ and $\lambda$ denote the weights for different losses. ${L}_{Total}$ is differentiable and the proposed network is trained in an end-to-end manner using a single view via unsupervised learning.

In the second stage, as illustrated in Fig. 2 (b), ${I}_{r}^{G}$ in a random pose ${{P}_{r}}$ is first synthesized from ${I}_{S}$ using the proposed model after the first stage training, then the proposed model uses ${I}_{r}^{G}$ as the source and ${I}_{S}$ as target to further train the network. This reverse mapping strategy has improved both  stability and performance of the network. Note that same loss function as shown in Eq. (4) is used in this stage.

%In the second stage, as depicted in Fig. 2 (b), firstly, taking the ${I}_{S}$ as the input of the TTM and VGM, the output view ${I}_{r}^{G}$ is obtained with a random pose ${{P}_{r}}$. In this way, the paired views with different poses, such as the output view ${I}_{r}^{G}$ and original input view ${I}_{S}$, are constructed for simulating the real process of novel view synthesis. After that, taking the ${I}_{r}^{G}$ as the input of the TTM and VGM, the view ${I}_{S}^{G}$ is synthesized at the pose ${{P}_{S}}$. In this way, the relative view transformation between ${I}_{r}^{G}$ and ${I}_{S}^{G}$ is learned to facilitate the unsupervised novel view synthesis. Note that, the same multiple losses with the first stage are utilized for training in the second stage.}}

%By utilizing the proposed reverse synthesis strategy, the transformation between input view and the output view is not limited. In other words, the pose of output view ${I}_{S}^{G}$ in reverse synthesis strategy is different from the pose of the input synthesized view ${I}_{r}^{G}$ during training phase, thus more view transformations are learned to a certain extent in a unsupervised manner.}}

\section{Experiments}
\subsection{Dataset and Implementation}

The proposed network is first verified on two popular categories, Chair and Car, of the ShapeNet dataset \cite{38}. There are 54 different camera poses for each object, 18 azimuth angles and 3 elevations. Due to limitation of GPU memory, input views are scaled to $160\times160\times3$. But for a fair comparison with the existing methods, output views are resized to $256\times256\times3$.  Same as \cite{2}, 80\% of the data are used for training, and the rest for testing. The commonly used ${{L}_{1}}$ distance and SSIM between the synthesized view and its ground-truth are adopted as the quantitative metrics.

The network is implemented using Pytorch framework \cite{39}, and Adam optimizer \cite{40} is adopted for training. The 2D encoder consists of five convolutional layers with 16, 32, 64, 128 and 256 filters of $3\times3$ and stride 2 each layer, respectively. The channel reduction/up-sampling block consists of five blocks of the convolutional layer (128, 64, 32, 16, and 3 filters of size $3\times3$, stride 1 and padding in each block) and up-sampling layers (scale 2). The 2D-3D encoder and 3D-2D decoder of the VGM are same as the ones in \cite{2}.

All experiments in this paper are conducted on a single GeForce GTX 1080Ti GPU with 11 GB of memory and Intel i7-8700K processor $@$3.70 GHz. The initial learning rate is set to 0.00005, and the batch size is set to 4. $\alpha$,  $\beta$, $\gamma$ and $\lambda$ are set empirically to 1, 5, 10 and 0.5, respectively. Additionally, models for each category are trained from scratch.  It took 6 days for Chair category, 14 days for Car category.

\subsection{Results and Comparison}

Table~\ref{table1} shows the performance of the proposed network and its comparison to other state-of-the-art methods including STM \cite{13}, AFF \cite{22}, MTN \cite{1} and TBN \cite{2}. Notice that these methods are trained using paired views with different poses though they synthesize novel view from a single source view with pose as the required auxiliary information.

\begin{figure}[!]
\centering
\includegraphics[width=1\linewidth]{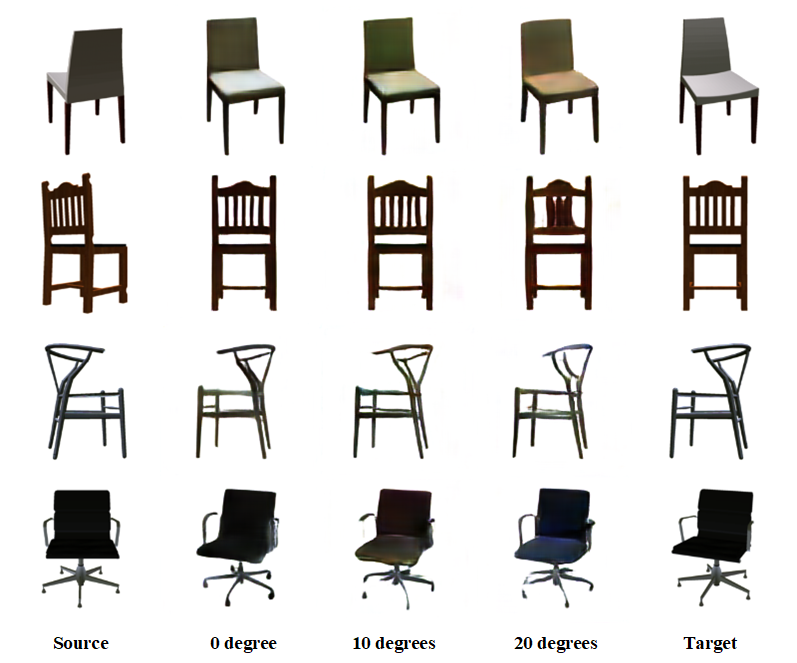}
\setlength{\abovecaptionskip}{-0.25cm}
\caption{Synthesized novel views from different elevation angles.}
\label{figure5}
\end{figure}

As seen from Table~\ref{table1}, the proposed network achieves comparative performance against the state-of-the-art methods under the condition of input source pose without gaussian noise. In particular, the proposed method has also made some improvements for object Chair, for example, the proposed method improves ${{L}_{1}}$ distance by 0.105, 0.091 and SSIM by 0.009 and 0.008 compared with the STM method and AFF method, respectively. This demonstrates the effectiveness of the proposed method for synthesizing a novel view from a single viewpoint image without pose, even though the network is trained using a single view, rather than paired views with different poses as these two methods do. In addition, the proposed method utilizes 3D representation to perform view transformation. As for the comparison with MTN method and TBN method, the proposed method improves ${{L}_{1}}$ distance by 0.017 and 0.014 respectively, and achieves comparable SSIM. We further note that the proposed method achieves comparable results to the state-of-the art methods for object Car, as shown in Table~\ref{table1}.

\begin{figure}[!]
\centering
\includegraphics[width=1\linewidth]{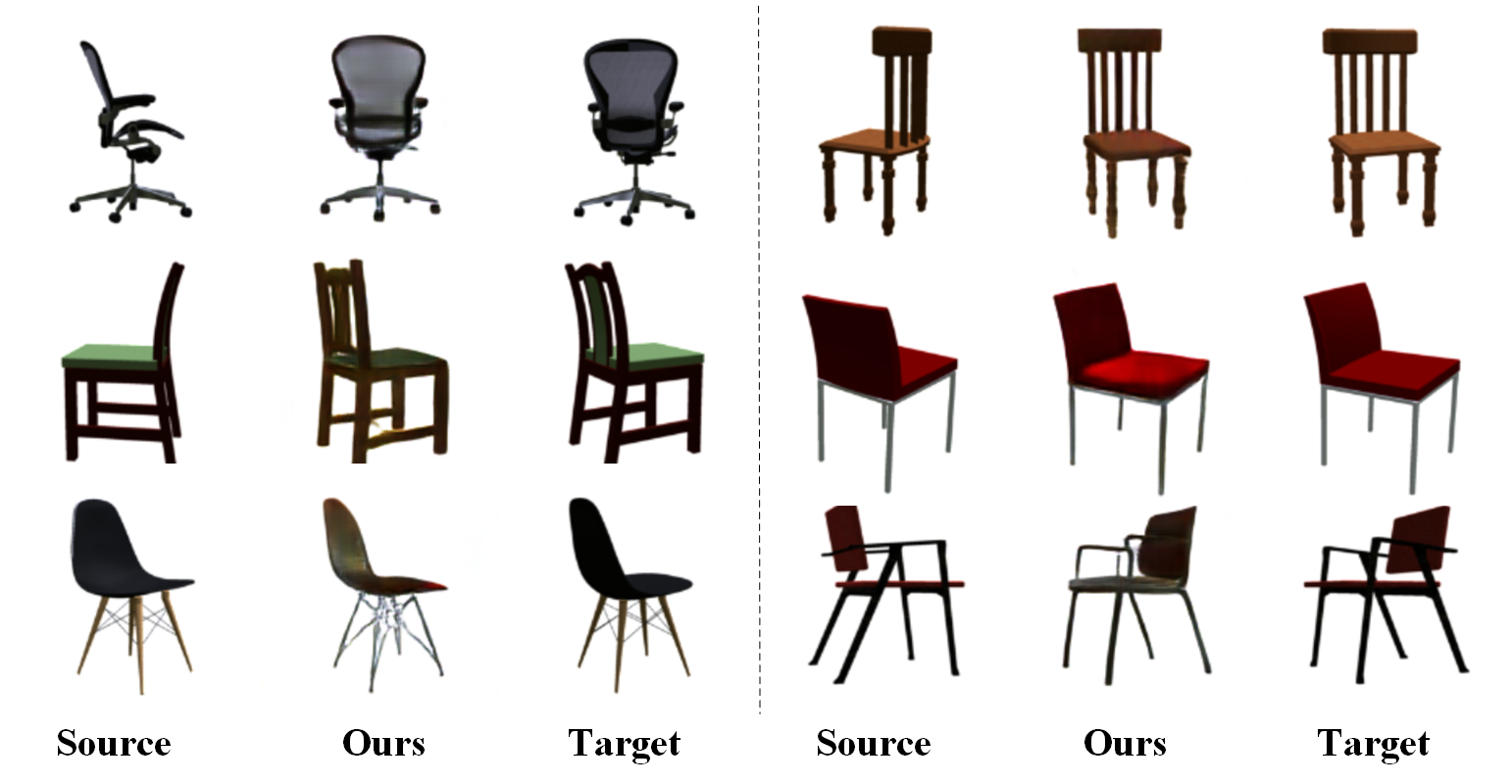}
\setlength{\abovecaptionskip}{-0.25cm}
\caption{Synthesis from an arbitrary image on Chair category.}
\label{figure6}
\end{figure}

\begin{figure}[!]
\centering
\includegraphics[width=1\linewidth]{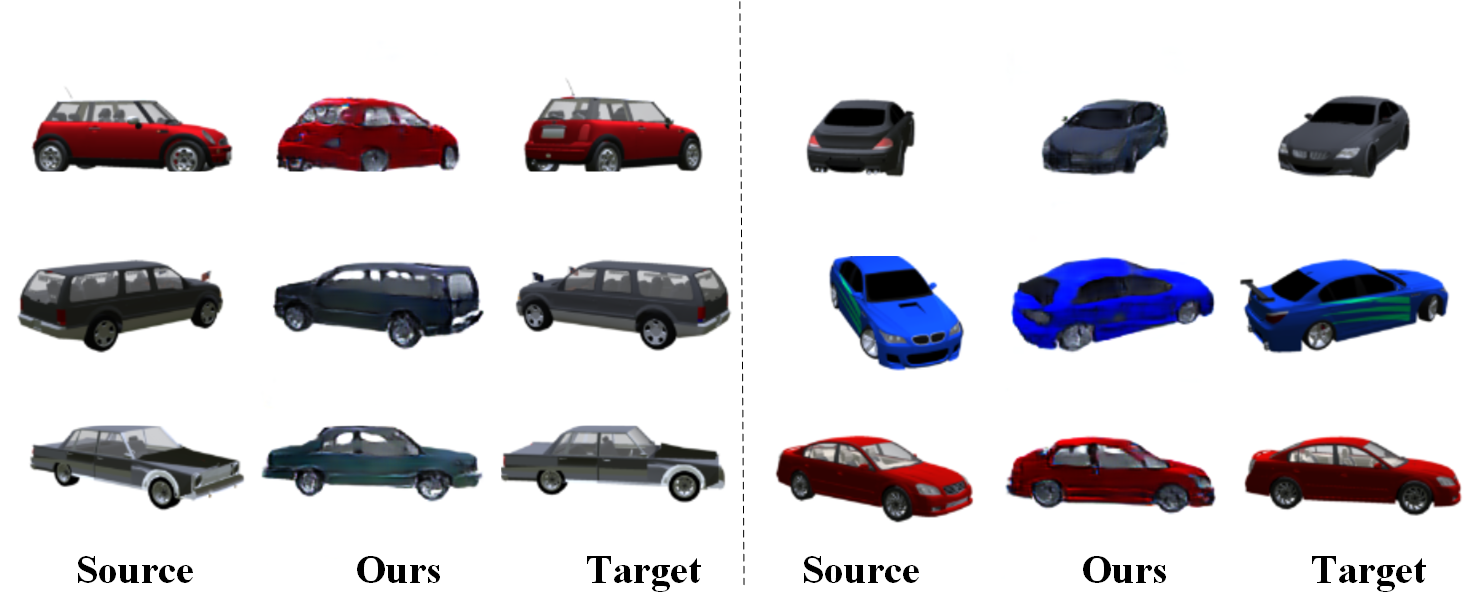}
\setlength{\abovecaptionskip}{-0.25cm}
\caption{Synthesis from an arbitrary image on Car category.}
\label{figure7}
\end{figure}

Fig.~\ref{figure2} and Fig.~\ref{figure3} show the visual comparisons of the novel views synthesized by the proposed method and compared methods under the condition of input source pose without Gaussian noise. As seen, the proposed method can generate a sharp novel view. In contrast, views synthesized by STM fail to generate much texture. Although the views synthesized by AFF preserve texture, it fails to generate pixels in some areas. Importantly, the views generated by the proposed method are relatively sharp mainly contributed by the feature transformation and 3D volumetric representation. The views generated by the proposed method are comparable to the views generated by MTN and TBN.

In addition, to investigate impact of noise in the pose of the input source view to the quality of novel views, small Gaussian noise with zero mean and standard deviation 1 is added to the source pose during the synthesis. Since the proposed method does not require pose information of input source view for synthesis, there is not impact to its performance. However, for the existing methods, such as STM \cite{13}, AFF \cite{22} and MTN \cite{1}, that require accuracy pose information of input source view for synthesis, it is found that the quality of synthesized views has been degraded as shown in Table~\ref{table1} where the noise column is ticked.

\begin{table}[t]
\newcommand{\tabincell}[2]{\begin{tabular}{@{}#1@{}}#2\end{tabular}}
\renewcommand\arraystretch{1.3}
\centering
\small
\caption{Quality of novel views of Chair synthesized by the proposed method with different ${P}_{R}$ (azi, ele).}
\begin{tabular}{p{3cm}<{\centering}p{2cm}<{\centering}p{2cm}<{\centering}}
\toprule[1.5pt]
    \multirow{2}[1]{*}{(azi, ele)} & \multicolumn{2}{c}{Chair} \\
            & \multirow{1}[0]{*}{${{L}_{1}}$} & \multirow{1}[0]{*}{SSIM} \\
    \midrule
     (0, 0)  & 0.164  & 0.879 \\
     (30, 0) & 0.177  & 0.873 \\
     (60, 0) & 0.167  & 0.877 \\
     (90, 0) & 0.175  & 0.873 \\
     (120, 0) & 0.173  & 0.875 \\
     (150, 0) & 0.169  & 0.876 \\
     (180, 0) & 0.174  & 0.873 \\
\bottomrule[1.2pt]
\end{tabular}

\label{table3}
\end{table}

\begin{table}[t]
\newcommand{\tabincell}[3]{\begin{tabular}{@{}#1@{}}#2\end{tabular}}
\renewcommand\arraystretch{1.3}
\centering
\small
\caption{Quality of novel views of Chair synthesized by the proposed method with different ${P}_{R}$ (azi, ele).}
\begin{tabular}{p{3cm}<{\centering}p{2cm}<{\centering}p{2cm}<{\centering}}
\toprule[1.5pt]
    \multirow{2}[1]{*}{(azi, ele)} & \multicolumn{2}{c}{Chair} \\
            & \multirow{1}[0]{*}{${{L}_{1}}$} & \multirow{1}[0]{*}{SSIM} \\
    \midrule
     (0, 0)  & 0.164  & 0.879 \\
     (0, 10) & 0.171  & 0.875 \\
     (0, 20) & 0.174  & 0.874\\
\bottomrule[1.2pt]
\end{tabular}

\label{table4}
\end{table}

\begin{table}[t]
\newcommand{\tabincell}[4]{\begin{tabular}{@{}#1@{}}#2\end{tabular}}
\renewcommand\arraystretch{1.3}
\centering
\small
\caption{Synthesis from an arbitrary image.}
\begin{tabular}{p{3cm}<{\centering}p{2cm}<{\centering}p{2cm}<{\centering}}
\toprule[1.5pt]
    \multirow{1}[1]{*}{Category} & \multirow{1}[1]{*}{${{L}_{1}}$} & \multirow{1}[1]{*}{SSIM} \\
    \midrule
     Chair  & 0.168  & 0.877 \\
     Car    & 0.137  & 0.888 \\
\bottomrule[1.2pt]
\end{tabular}

\label{table5}
\end{table}

\subsection{Impact of Choice of Reference Pose ${{P}_{R}}$}
In this subsection, the effect of the choice of reference pose ${{P}_{R}}$ is studied. Experiments are conducted on the object Chair by selecting different ${{P}_{R}}$. Considering object Chair is symmetrical, 10 different ${{P}_{R}}$ are sampled whose azimuth angle ranges from 0 to 180 degrees with an interval of 30 degrees and elevation angle ranges from 0 to 20 degrees with an interval of 10 degrees.

Results of some selected ${{P}_{R}}$ are shown in Table~\ref{table3} and Table~\ref{table4}, where ``azi'' represents azimuth and ``ele'' represents elevation. It can be seen both indicators ${{L}_{1}}$ distance and SSIM, do not vary much among these different reference poses. This demonstrates that a novel view is synthesized by rotating the explicit occupancy volume with respect to the view of arbitrary selected reference poses. Reference pose ${{P}_{R}}(0,0)$ achieves the best indicators. This is because view images of object Chair are sampled from its CAD model, their corresponding viewpoints are set relative to 0 degree azimuth angle and 0 degree elevation angle in the ShapeNet dataset.

For visual inspection, multiple novel views synthesized by the proposed method with different reference poses are shown in Fig.~\ref{figure4} and Fig.~\ref{figure5}. Fig.~\ref{figure4} shows the results with different azimuth angles and 0 degree elevation angle. Fig.~\ref{figure5} shows the results with different elevation angles and 0 degree azimuth angle. It can be seen that all novel views are well-synthesized despite different reference poses. This demonstrates the expectation of the design using a reference pose ${{P}_{R}}$.
\begin{figure*}[!t]
\centering
\includegraphics[width=0.8\linewidth]{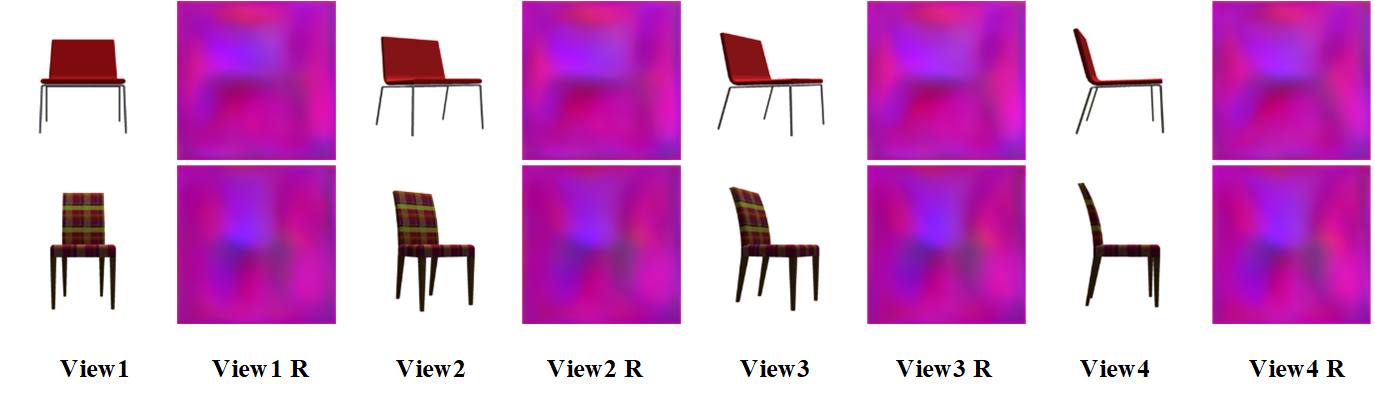}
\setlength{\abovecaptionskip}{-0.25cm}
\caption{Visualization of the features from multiple different views in TTM.}
\label{figure8}
\end{figure*}

\begin{figure*}[!t]
\centering
\includegraphics[width=0.8\linewidth]{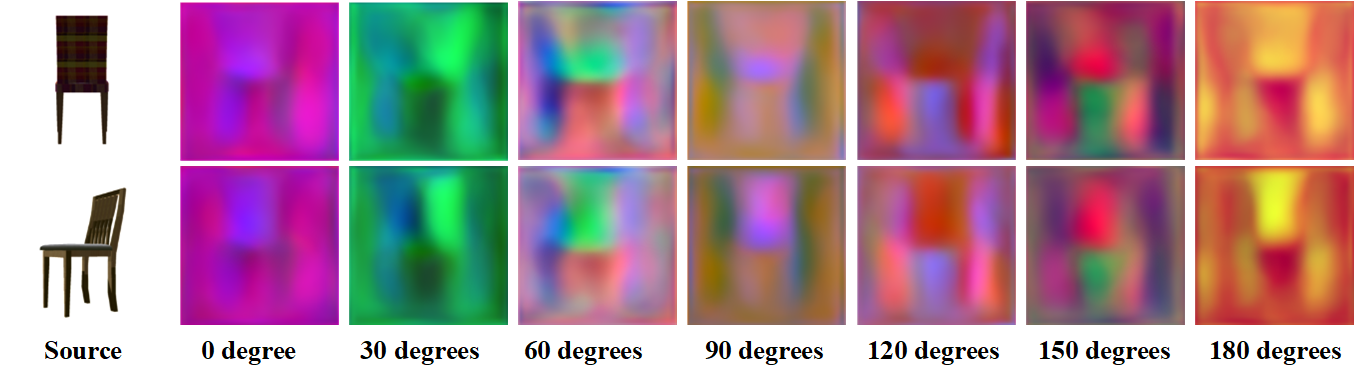}
\setlength{\abovecaptionskip}{-0.25cm}
\caption{Visualization of the features with different azimuth angles in TTM.}
\label{figure9}
\end{figure*}

\subsection{Synthesis from an Arbitrary Single Image}
In this experiment, we simulates a practical case to synthesize novel views from an image captured by an arbitrary camera. We choose source views in synthesis whose viewpoints are not included in training. In other words, these viewpoints using in synthesis are new to the trained network. Results are shown in Table~\ref{table5}, the quantitative results show the proposed method can effectively generate a novel view from an arbitrary image. Fig.~\ref{figure6} and Fig.~\ref{figure7} show some examples.

\begin{table}[!t]
\newcommand{\tabincell}[2]{\begin{tabular}{@{}#1@{}}#2\end{tabular}}
\renewcommand\arraystretch{1.3}
\centering
\small
\caption{Evaluation of the reserve training strategy on Chair.}
\begin{tabular}{p{5.8cm}<{\centering}p{0.8cm}<{\centering}p{0.8cm}<{\centering}}
\toprule[1.2pt]
\multirow{2}{*}{Training Method}
& \multicolumn{2}{c}{Chair}\\
%& Sketchy &  \tabincell{c}{TU_Berlin\\Extension}\\
& ${{L}_{1}}$& SSIM \\
\cline{1-3}
\hline
Without reverse mapping strategy     & $0.174$ & $0.878$ \\
\textbf{With reverse mapping strategy}   & $\textbf{0.164}$ & $\textbf{0.879}$ \\
\bottomrule[1.2pt]
\end{tabular}
\label{table6}
\end{table}

\begin{figure}[!t]
\centering
\includegraphics[width=0.8\linewidth]{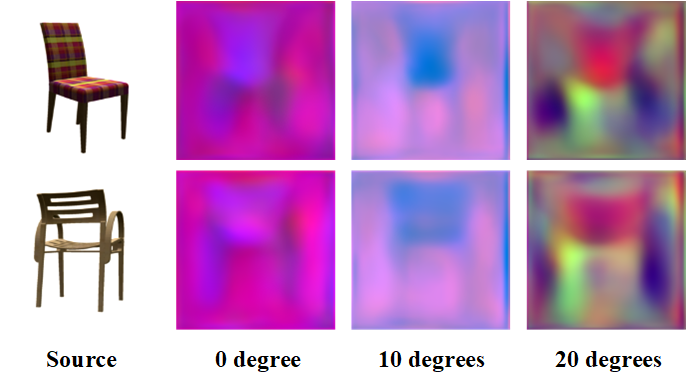}
\setlength{\abovecaptionskip}{-0.25cm}
\caption{Visualization of the features with different elevation angles in TTM.}
\label{figure10}
\end{figure}

\subsection{Analysis of Intrinsic Representation in TTM}
In this subsection, the visualization analysis of the intrinsic representation in TTM is studied. With respect to the reference pose (i.e. 0 degree azimuth angle and 0 degree elevation angle), the intrinsic representations from different viewpoint images of the same object are visualized in Fig.~\ref{figure8}. As seen, with different viewpoint images as input, even though there are large azimuth angles and elevation angles across different views, the produced intrinsic representations have not changed much. This demonstrates the expected design that TTM produces an intrinsic representation with respect to a specific pose, that is the pre-defined reference pose in the proposed method.

Besides, multiple intrinsic representations generated by the proposed method with different reference poses are shown in Fig.~\ref{figure9} and Fig.~\ref{figure10}. Fig.~\ref{figure9} shows the results with different azimuth angles and 0 degree elevation angle. Fig.~\ref{figure10} shows the results with different elevation angles and 0 degree azimuth angle. It can be seen that the intrinsic representations changes with the rotation angle, but the representations with the same rotation angle from different input source views are similar. It proves that the proposed TTM can transform the input source viewpoint image to obtain a intrinsic representation at the reference pose.

\begin{figure}[!t]
\centering
\includegraphics[width=0.8\linewidth]{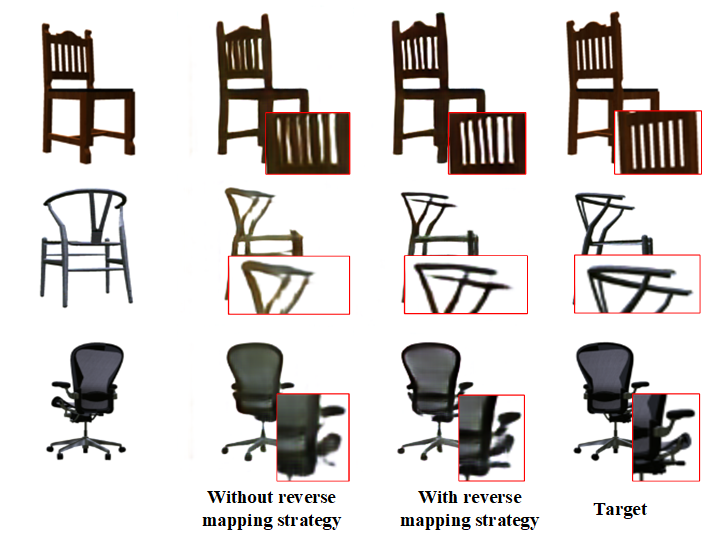}
\setlength{\abovecaptionskip}{-0.25cm}
\caption{Visualization of some synthesized views with and without reverse mapping training.}
\label{figure11}
\end{figure}

\subsection{Contribution of the Reverse Mapping Training Strategy}
In order to demonstrate the effectiveness of the reverse mapping strategy, experiments are conducted to compare the performance of the proposed network with and without this training strategy. Results are shown in Table V. As seen, the reverse mapping strategy improves both ${{L}_{1}}$ distance and SSIM. Fig. 12 compares visually the synthesized views with and without reverse mapping strategy. As seen, the training of reverse mapping strategy improves the capability of the model to generate views with more accurate appearance and structure compared to those generated by the model trained without the reverse mapping strategy.

\section{Conclusion}

This paper presents an unsupervised network for synthesizing a novel view from a single image without requiring pose information of the source view. With the support of a specifically designed token transformation module (TTM), a view generation module (VGM), and a reverse mapping strategy, the network is trained with a single view in an unsupervised manner. The network facilities a processing pipeline of feature transformation to a reference pose, reconstruction of 3D volumetric representation, and rendering of the 3D volume from a novel pose. One of the key advantages is that the proposed network enables a new feature in a multi-view system, that is, generating a novel view from any source viewpoint images capturing by any camera that are not part of multi-view source cameras. It is expected that there is a canonical view for any object from which the view of the most representative. Instead of setting the reference pose ${{P}_{R}}$, an optimal ${{P}_{R}}$ can and should be learned as well. Also, for a scene, multiple ${{P}_{R}}$ may be required to cover the entire scene. Both the optimal and multiple reference poses will be studied in the future.

\ifCLASSOPTIONcaptionsoff
  \newpage
\fi

\end{document}